\documentclass[letterpaper, 10 pt, conference]{ieeeconf}  

\usepackage{graphicx}
\graphicspath{{images/}}
\usepackage{amsmath}
\usepackage{subcaption}
\usepackage[thinlines]{easytable}
\usepackage{cite}
\usepackage{float}

\IEEEoverridecommandlockouts                              

\usepackage{color}
\usepackage{booktabs}
\usepackage{textcomp}
\usepackage{microtype} 
\usepackage{hyperref}
\hypersetup{
    colorlinks=true,
    linkcolor=blue,
    }
\usepackage{ccicons}  

\usepackage{xspace}

\usepackage{tabularx}
\usepackage[labelfont=bf,font=footnotesize]{caption}
\usepackage{amssymb}
\usepackage{multirow}
\usepackage{verbatim}
\usepackage{booktabs}

\renewcommand{\baselinestretch}{0.97}

\makeatletter
\DeclareRobustCommand\onedot{\futurelet\@let@token\@onedot}
\def\@onedot{\ifx\@let@token.\else.\null\fi\xspace}

\makeatother

\usepackage{lipsum}

\usepackage[dvipsnames]{xcolor}     
\usepackage{amsfonts}               
\usepackage{mathrsfs}
\let\labelindent\relax             
\usepackage{enumitem}              
\usepackage{algorithm}             
\usepackage[noend]{algpseudocode}  

\usepackage{cuted}

\long\def\xspace{\mathcal{X}}

\long\def\sl#1{\textcolor{orange}{\bf \small [SL: #1d]}}

\title{\LARGE \bf
MBot: A Modular Ecosystem for Scalable Robotics Education
}

\author{
    Peter Gaskell 
    \and Jana Pavlasek 
    \and Tom Gao
    \and Abhishek Narula
    \and Stanley Lewis
    \and Odest Chadwicke Jenkins
\thanks{Authors are with the Robotics Institute and Robotics Department, University of Michigan, Ann Arbor, USA 48109-2106. 
        {\tt\footnotesize [pgaskell, pavlasek, zimingg, abnarula, stanlew, ocj]@umich.edu}.}
}

\begin{document}

\maketitle
\thispagestyle{empty}
\pagestyle{empty}


\begin{strip}\centering
\vspace{-1.5cm}
    \includegraphics[width=\linewidth]{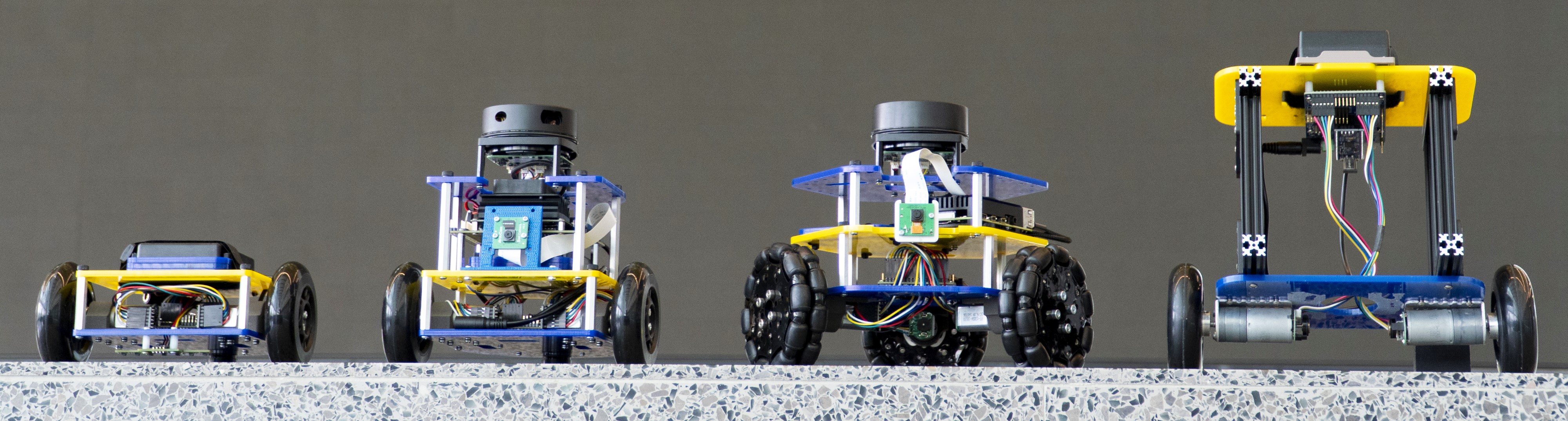}
    \label{fig:pitch}
\end{strip}

\begin{abstract}
The Michigan Robotics MBot is a low-cost mobile robot platform that has been used to train over 1,400 students in autonomous navigation since 2014 at the University of Michigan and our collaborating colleges.  The MBot platform was designed to meet the needs of teaching robotics at scale to match the growth of robotics as a field and an academic discipline.  Transformative advancements in robot navigation over the past decades have led to a significant demand for skilled roboticists across industry and academia. This demand has sparked a need for robotics courses in higher education, spanning all levels of undergraduate and graduate experiences. Incorporating real robot platforms into such courses and curricula is effective for conveying the unique challenges of programming embodied agents in real-world environments and sparking student interest. However, teaching with real robots remains challenging due to the cost of hardware and the development effort involved in adapting existing hardware for a new course. In this paper, we describe the design and evolution of the MBot platform, and the underlying principals of scalability and flexibility which are keys to its success. 

\end{abstract}

\section{Introduction}

The growing popularity of robotics and AI has led to an increasing demand for new courses which integrate cutting-edge concepts at all levels of undergraduate and graduate studies.  
And what is a robotics course without real robots?
The use of real robot platforms in robotics curricula has the potential to motivate and inspire students, while providing practical grounding for the course concepts.
However, many commercially available platforms are prohibitively expensive and lack the flexibility to support the diverse needs of students and instructors.
Open-source projects are often more customizable and affordable, but require a significant upfront development effort and consistent maintenance, making their integration into curricula impractical for many institutions.

The \textbf{MBot} is a low-cost, flexible platform for robotics education at the undergraduate and graduate levels designed at the University of Michigan.
The MBot is designed with modularity in mind, and includes a number of possible robot configurations.
The platform is the result of over a decade of teaching robotics and has served six courses and over 1,400 students since 2014.
The platform has been used to teach courses on robot localization, control, planning, and introductory programming, from the first year of undergraduate study through graduate-level courses.
The MBot has also been used as part of a distributed teaching initiative to offer robotics courses at collaborating institutions.

The MBot can be constructed from commercial, off-the-shelf components.
The platform is driven by a simple, versatile backbone, called the \textit{MBot Robotics Control Board}.  
The board can be used in a number of distinct configurations, from a low-cost version with basic sensing capabilities to a platform capable of advanced autonomy using a 2D Lidar and an RGB camera. 
The flexible design enables a configurable chassis and sensor suite with minimal core hardware modifications.

Central to the MBot's versatility as an educational platform is a comprehensive set of open-source software tools and a custom API for high-level programming, making it suitable for use in both advanced and introductory courses.
The robot can be programmed through multiple modes depending on the needs of the user and on the application. The software supports advanced autonomy applications such as mapping, localization, and perception through message-passing frameworks. Alternatively, users can program the robot using the custom \textit{MBot Bridge API}, which provides a simple, synchronous interface to the autonomy processes. 
The MBot platform also features a custom web application for remote control and visualization, which can be accessed from any personal computer. 

The MBot has built on this design to realize a low-cost mobile robot platform that has been used to train
students in robotics and AI
at the University of Michigan and our collaborating colleges.  The MBot platform was designed for seamless adoption into higher education curricula.  Specifically, the MBot aims to meet the needs of teaching robotics at scale to match the growth of robotics as a field and an academic discipline~\cite{jenkins2023michigan}, spanning all levels of undergraduate and graduate experiences. More information on the MBot platform is available at: \href{https://mbot.robotics.umich.edu/}{https://mbot.robotics.umich.edu/}.

\section{The Development of the MBot}\label{sec:history}
\begin{figure}
    \centering
    \includegraphics[width=0.95\linewidth]{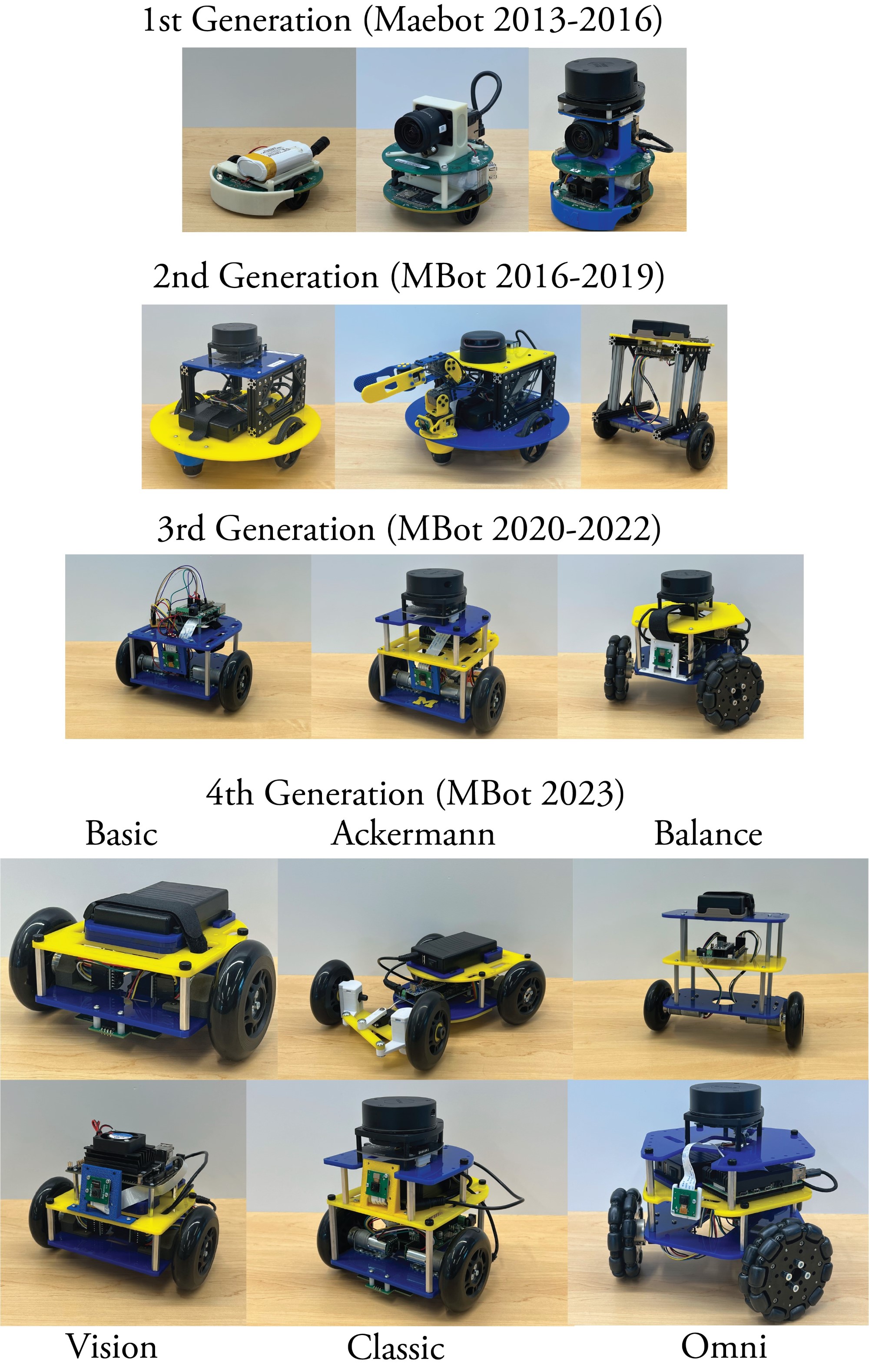}
    \caption{Ten years of teaching with the MBot.}
    \label{fig:mbot_history}
\end{figure}

The MBot has grown into an ecosystem of platforms and tools suitable for instruction of a variety of courses in higher education in response to lessons learned over nearly a decade of teaching. Figure~\ref{fig:mbot_history} shows the evolution of the platform.
The Maebot platform \cite{maebot} was originally developed for instruction of a final-year undergraduate course on autonomous robotics at the University of Michigan in 2014. In 2016, the MBot was designed as a lower-cost version of the original design and expanded to teach the graduate-level course \textit{Robotic Systems Lab} as part of the University of Michigan's graduate degree in robotics. 

Early generations of the platform focused on affordability and the incorporation of features suitable for supporting the learning objectives for advanced robotics courses, such as a 2D Lidar for mapping and navigation. 
The COVID-19 pandemic sheltering necessitated a new generation of the platform with enhanced tools suitable for remote learning efforts.
Students in robotics lab classes were able to successfully achieve course learning objectives in their home environments with typical compute resources when provided only with a basic MBot kit of parts shipped to them and remote support by the teaching staff.

\begin{figure}
    \centering
    \includegraphics[width=\linewidth]{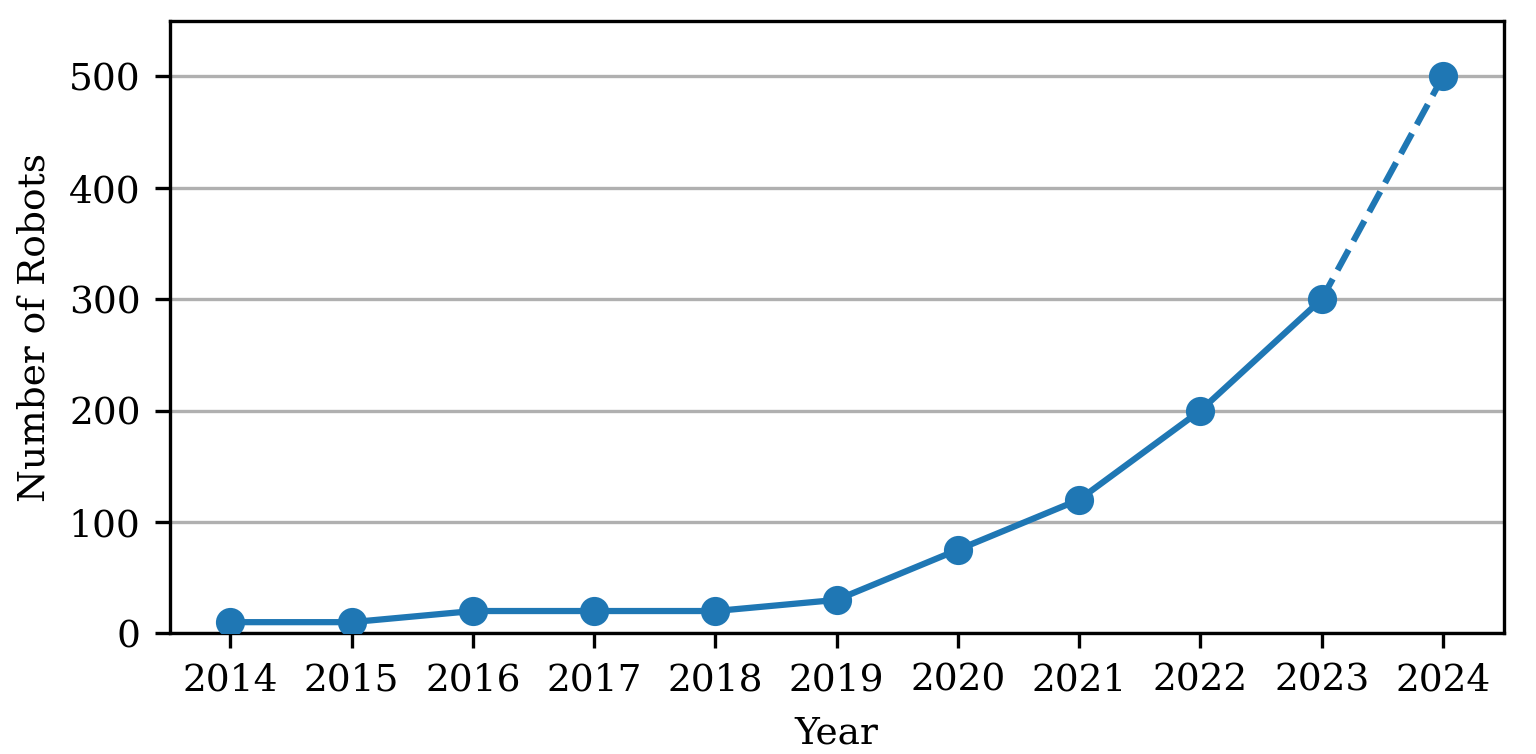}
    \caption{Growth of MBot fleet over time to meet teaching needs for robotics classes at Michigan. The dotted line represents projected numbers.}
    \label{fig:mbot_adoption}
\end{figure}

The founding of the University of Michigan Robotics undergraduate degree~\cite{jenkins2023michigan} in 2022 expanded the number of robotics courses offered, introducing a number of new courses with hands-on components.
Growing enrollment introduced new constraints on the MBot platform, including that the platform support courses offered early in the degree with minimal or no programming prerequisites.
This sparked the development of the latest generation of the MBot family of platforms which includes an ecosystem of tools for interacting with the robot at different levels, including a custom web application for visualization and an API.
Multiple additional configurations of the robot, including a low-cost budget version and an omnidrive platform, were also introduced as part of the expanded robotics curriculum. Figure~\ref{fig:mbot_adoption} shows the growth of the MBot fleet over time. 

The MBot has also been used as part of a \textit{distributed teaching collaborative} between institutions.
Building off of lessons learned during the pandemic when robots were used in students' homes, the MBots have been deployed at Berea College, Howard University, and Morehouse College to teaching autonomous robotics to undergraduate students.

\section{The MBot Platform}\label{sec:mbot}

\begin{table}[t]
\centering
\caption{\label{tab:mbot_bom} MBot Bill of Materials}
\renewcommand{\arraystretch}{1.2}
\small
\begin{tabular}{r c c c c c}
\toprule
& & \multicolumn{3}{c}{\textbf{MBot Version}} \\
 & Approx. &  & Classic & Classic & Omni \\ 
Component & Cost & Basic & RPi & Jetson & RPi \\ 
\midrule
Control board & \$15 & $\checkmark$ & $\checkmark$ & $\checkmark$ & $\checkmark$\\
Pico & \$5 & $\checkmark$ & $\checkmark$ & $\checkmark$ & $\checkmark$ \\
Motors & \$15 & $\checkmark$ & $\checkmark$ & $\checkmark$ & $\checkmark$ \\
Battery & \$25 & $\checkmark$ & $\checkmark$ & $\checkmark$  & $\checkmark$ \\
Basic chassis & \$20 & $\checkmark$ & $\checkmark$ & $\checkmark$  &  \\
Omni chassis & \$40 &  &  &  & $\checkmark$ \\
Basic wheels& \$20 & $\checkmark$ & $\checkmark$ & $\checkmark$ & \\
Omni wheels & \$80 & & & & $\checkmark$ \\
RPi 4 kit & \$85 & & $\checkmark$ & & $\checkmark$ \\
Jetson kit & \$175 & & & $\checkmark$ & \\
Lidar & \$100 &  & $\checkmark$ & $\checkmark$ & $\checkmark$ \\
\midrule
\textbf{Total:} & & \$100 & \$285 & \$375 & \$385 \\
\bottomrule
\end{tabular}
\end{table}

The guiding principles for the design of the platform are as follows:
\begin{itemize}
    \item a low-cost basic version, 
    \item low barrier to entry for students and instructors,
    \item a modular, repairable, configurable \& extensible design,
    \item compatibility with advanced robotics tools (e.g. ROS~\cite{quigley2009ros, macenski2022robot}) for graduate education \& research.
\end{itemize}
The cost of the three most common versions of the platform (the Basic, the Classic, and the Omni) can be found in Table~\ref{tab:mbot_bom}. The hardware design is described below.

\subsection{Configurable Hardware Design}

The Basic version of the MBot platform forms the foundation for all platforms in the MBot family. It is centered around the MBot control board and can use either 12V or 6V 20mm motors and comes complete with a chassis, motor mounts, wheels and magnetic encoders. The Basic model is powered by a 12V 3Ah Li-Ion battery pack, and uses a 7-segment line sensor for line following and navigational tasks.  The BOM cost for the Basic model is under \$100 USD as of Fall 2023. The chassis plates have mounting holes for adding additional sensors, and creating a new chassis can be done quickly with a laser cutter. Building on the Basic model, the Balance add-on provides the parts to turn the Basic model into a mobile inverted pendulum robot and includes a redesigned chassis, additional hardware, and larger 12V 25mm motors for enhanced speed and torque.  

The Classic version is geared towards advanced robotic applications, specifically around visual processing, mapping, and navigation. It includes a single-board computer as its high-level computing module. Currently supported computing boards include the Raspberry Pi 4B+ or the NVidia Jetson Nano. The platform also supports additional sensors, including a 2D Lidar module to enable mapping and localization, and a camera allows for vision tasks such as visual odometry and object detection.  The Classic can be configured with different drive types, including Omnidrive and Ackermann versions.

\subsection{MBot Control Board}

\begin{figure}
    \centering
    \includegraphics[width=0.9\linewidth]{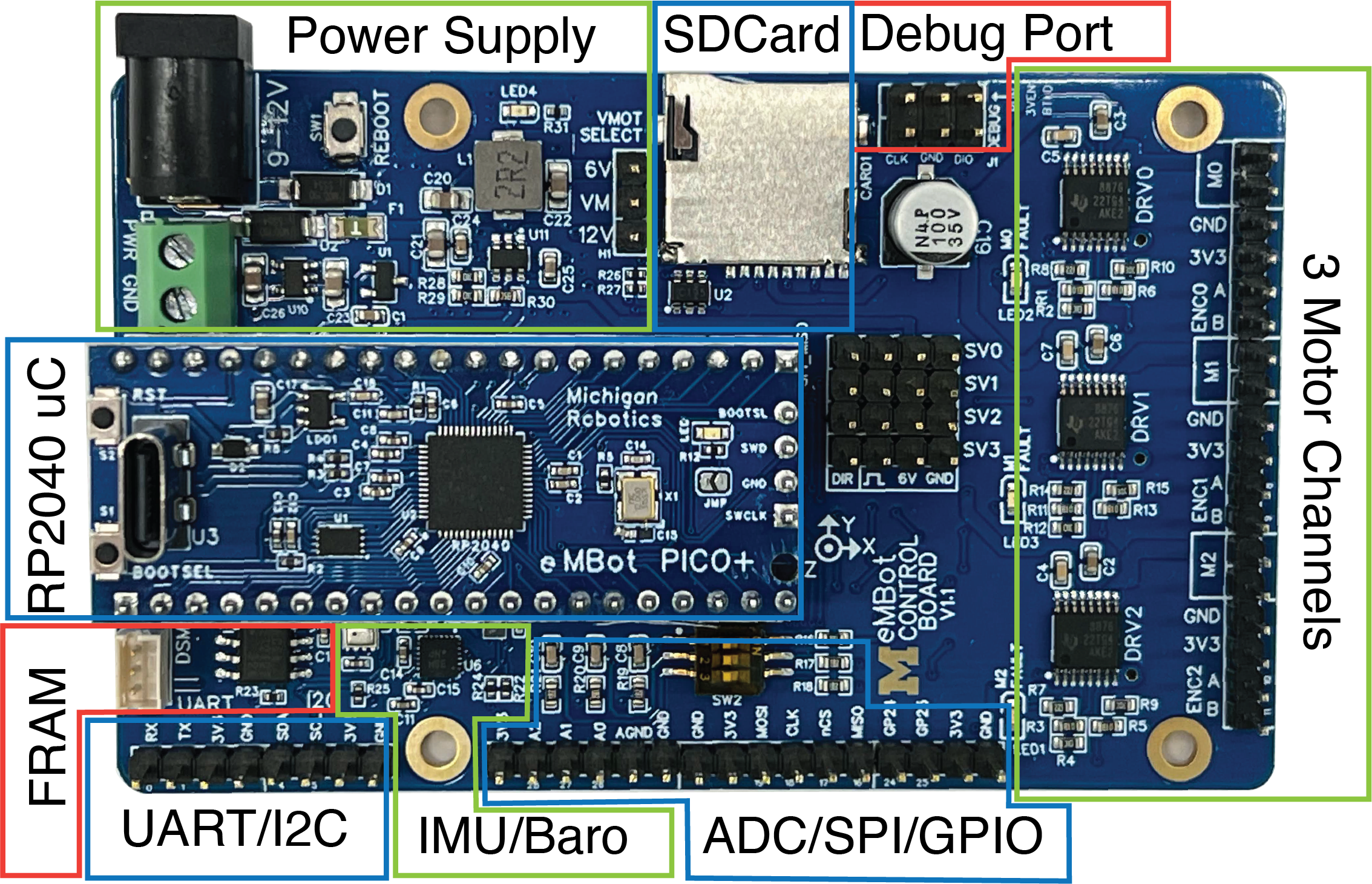}
    \caption{Image of the Robotics Control Board and custom RP2040 microcontroller based off the Raspberry Pi Pico.  The subsystems of the board are labeled. Each motor channel can drive a brushed DC motor and read signals from a quadrature encoder.  A forth motor driver can be added by attaching an external module.}
    \label{fig:ctrl_board}
\end{figure}

The foundation of the MBot platform is the MBot Control Board, shown in Figure~\ref{fig:ctrl_board}.  The board uses a RP2040 based microcontroller, and is compatible with the Raspberry Pi Pico family of microcontroller modules. It allows for control of three brushed DC motors with current feedback and relative position feedback using a quadrature encoder.  Each motor driver can deliver 1.3A continuous current.  The motor voltage can be selected by an onboard jumper between the applied DC voltage and the onboard 6V, 4A voltage regulator.  The control signals for the 3 motors, along with two additional motor control signals are broken out so the board can be used to drive four external motor drivers.  Any of the the motor channels can instead be configured as a hobby servo control signal to drive up to 4 servo motors.

The board contains various sensors and storage capabilities. A 9 degree of freedom MEMS Inertial Measurement Unit with  gyroscope, accelerometer, and magnetometer components performs data fusion to yield pose estimates. A barometer is also available to estimate altitude relative to an initial pressure reading.  A nonvolatile memory chip is included to store calibration data and parameters onboard the controller, and a SD card connector enables data logging without a host computer.

To connect external sensors and other devices, the board has dedicated ports for communication over common serial protocols including I2C and SPI.  The three analog to digital converters can be switched from measuring the current draw of the motors to measuring an external voltage.  Additionally, the debug pins near the card reader allows for soft/hard resets of the board, loading of different programs onto the board, and software debugging connections to enable debugging tools such as the GNU Debugger.

\begin{figure*}
    \centering
    \includegraphics[width=\linewidth]{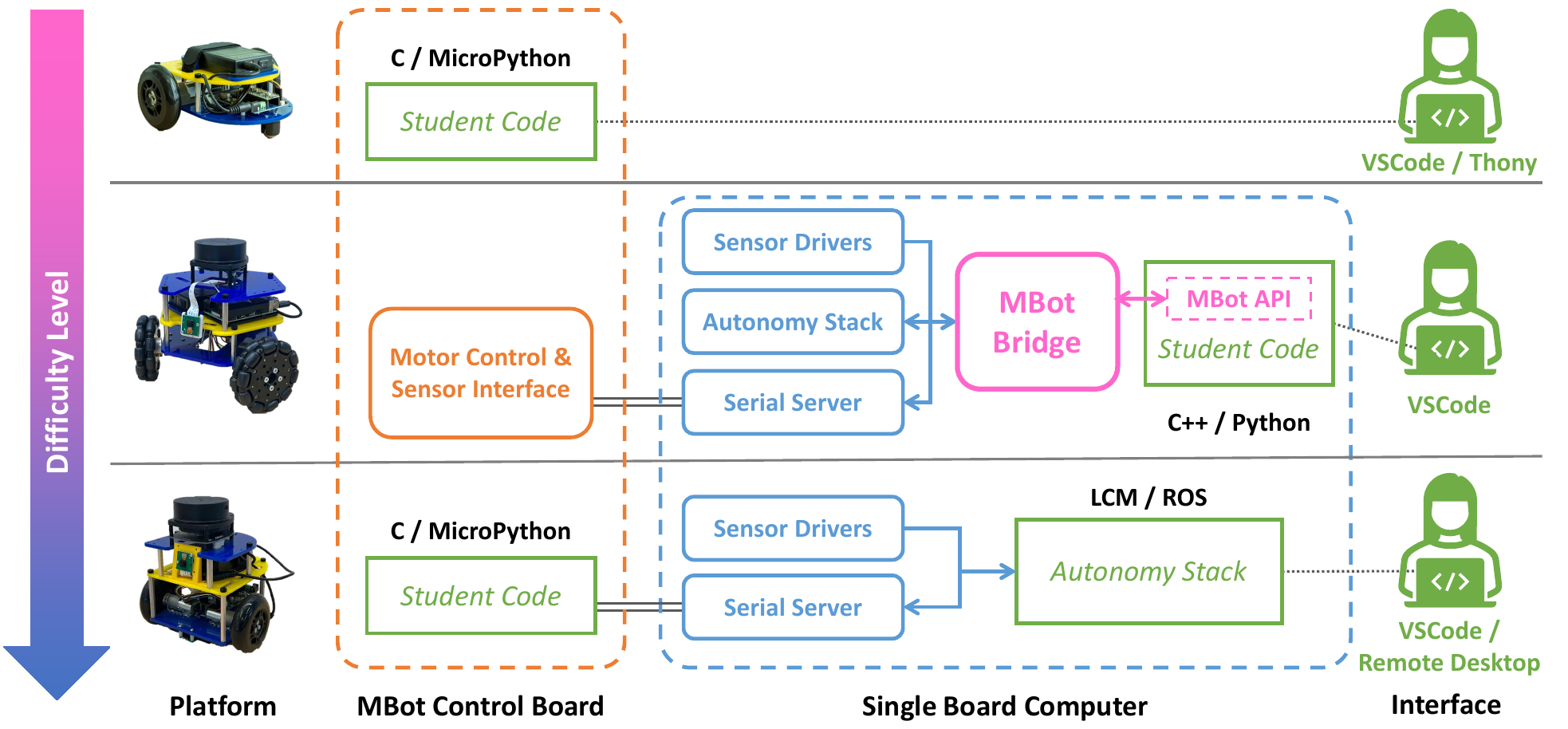}
    \caption{Possible software configurations for different levels of robotics courses. The Basic configuration with no single board computer can be programmed directly for basic applications using C or MicroPython (top). Classic and Omni versions can be programmed through the MBot Bridge API, which provides a simple interface for programming the MBot (middle). Advanced applications can interface directly with the message passing framework (bottom).}
    \label{fig:teaching_arch}
\end{figure*}

\subsection{MBot Firmware}

The MBot provides an accessible and user-friendly firmware solution for the RP2040 microcontroller. 
It offers firmware solutions in both C and MicroPython, enhancing the ease of implementing new MBot configurations, regardless of the user's programming or embedded development expertise. The clear separation between low-level interface and communication protocols and user-space code allows the robot to be customized for specific learning objectives without specialized, device-specific knowledge.
Meanwhile, experienced users can access the underlying driver codebase for further enhancements and modifications if desired and the community and support for the low-cost RP2040 microcontroller makes enhancing the codebase accessible. This distinct advantage positions the MBot Control Board as a preferred choice over alternative options such as ST Nucleo boards which require substantial embedded experience to master, or the higher cost Beaglebone Blue, which while a capable platform for robotics education~\cite{bewley2015leveraging}, is overpowered for simple motor control and sensor interfacing tasks.

The firmware of the MBot Control Board facilitates seamless communication with other devices through two virtual USB serial ports. One of these ports is dedicated to message-based communication with higher-level computing devices (e.g., a Raspberry Pi) using the ROSSerial protocol~\cite{bouchier2013embedded}. This capability enables the MBot Control Board to engage in communication using a publish/subscribe approach. When coupled with one of the serial communication service components of the MBot software ecosystem, the MBot Control Board can effortlessly publish and subscribe to messages used by higher-level computing devices. The second serial port is reserved for user-space code, allowing users to transmit human-readable data for purposes such as status reporting, ad-hoc data collection, or debugging.

Moreover, the MBot Control Board offers support for MicroROS \cite{belsareMicroROS2023}, enabling DDS communications with ROS2 systems. As ROS2 continues to gain prominence in the field of robotics research, this feature not only future-proofs the MBot Control Board but also enhances its compatibility with existing robotics systems. This forward-thinking approach ensures that the MBot remains a versatile and adaptable platform for robotics education and research.

\section{The MBot Architecture}\label{sec:mbot_software}

The MBot software architecture is designed to be both flexible and easy to use. 
The architecture enables users to select an interface based on the desired application and level of difficulty, making the platform suitable for introductory and advanced courses.
The MBot comes equipped with a full asynchronous software stack, including sensor drivers, mapping and localization, and path planning, in addition to an easy-to-use synchronous API. The robot also has a custom web app which can be used for visualization and basic control.
For advanced applications, the robot can be programmed by interfacing directly with the core software stack through an asynchronous message passing framework (e.g. LCM~\cite{huang2010lcm} or ROS~\cite{quigley2009ros, macenski2022robot}).
For single-threaded, synchronous applications, such as introductory programming assignments or quick prototyping, the \textit{MBot Bridge API} provides a simple interface to the MBot software stack. The software for various configurations is shown in Figure~\ref{fig:teaching_arch}.

The typical workflow for programming the MBot uses the single-board computer running a Linux-based operating system as the development environment.
Users connect to the remote computer using a remote session on a local IDE (e.g. VSCode's Remote extension), or through a remote desktop. 
This enables the robots to be programmed with minimal configuration on a personal computer.

In the following sections, we describe the functionalities available in the MBot software stack as well as two key tools we have developed for easy interaction with the MBots: the MBot Bridge API and the MBot Web App.

\subsection{The MBot Software Stack}

The software stack consists of drivers, utilities, and autonomy programs built on the asynchronous message-passing communication protocol Lightweight Communications and Marshalling (LCM)~\cite{huang2010lcm}. 
The available functionalities as of the time of writing include a driver for the Lidar, a serial interface to the MBot Control Board, a path tracker, and a SLAM node which employs Monte-Carlo Localization~\cite{probrob:thrun, gutmann2002experimental}. 
The SLAM implementation enables on-the-fly mode switching between full mapping and localization, localization only, and idle modes, which can be controlled through the web app.
We note that while the MBot software uses LCM, the platform is also compatible with other frameworks such as the Robot Operating System (ROS)~\cite{quigley2009ros, macenski2022robot}.

For single-threaded use cases, the MBot software can be used as a tool for developing downstream applications. For example, students can use the MBot's SLAM for mapping and localization through the web app, then implement an autonomous navigation algorithm by accessing the localization information through the API (see the middle row of Figure~\ref{fig:teaching_arch}).
For advanced use cases, any process in the MBot's autonomy stack can be disabled and replaced with custom code. For example, for educational modules on mapping and localization, students can implement their own SLAM algorithm by interfacing directly with LCM or ROS.

\subsection{MBot Bridge API}

The MBot supports synchronous programming in C++ and Python through the MBot Bridge API. The API provides a simple interface for reading robot data and sending control commands in single-threaded programs.
The API depends on the MBot Bridge Server, which manages incoming messages from the software stack and stores them in queues. The server exposes a websocket-based protocol using a custom JavaScript message definition inspired by ROS Bridge~\cite{crick2017rosbridge}. 
An added utility of the MBot Bridge server and API is that the websocket-based interface enables any device on the robot's network to communicate using the defined protocol. This means that the MBot Bridge can be used remotely from a personal laptop, and enables inter-robot communication within a robot fleet.

\subsection{MBot Web App}

\begin{figure}
    \centering
    \includegraphics[width=\linewidth]{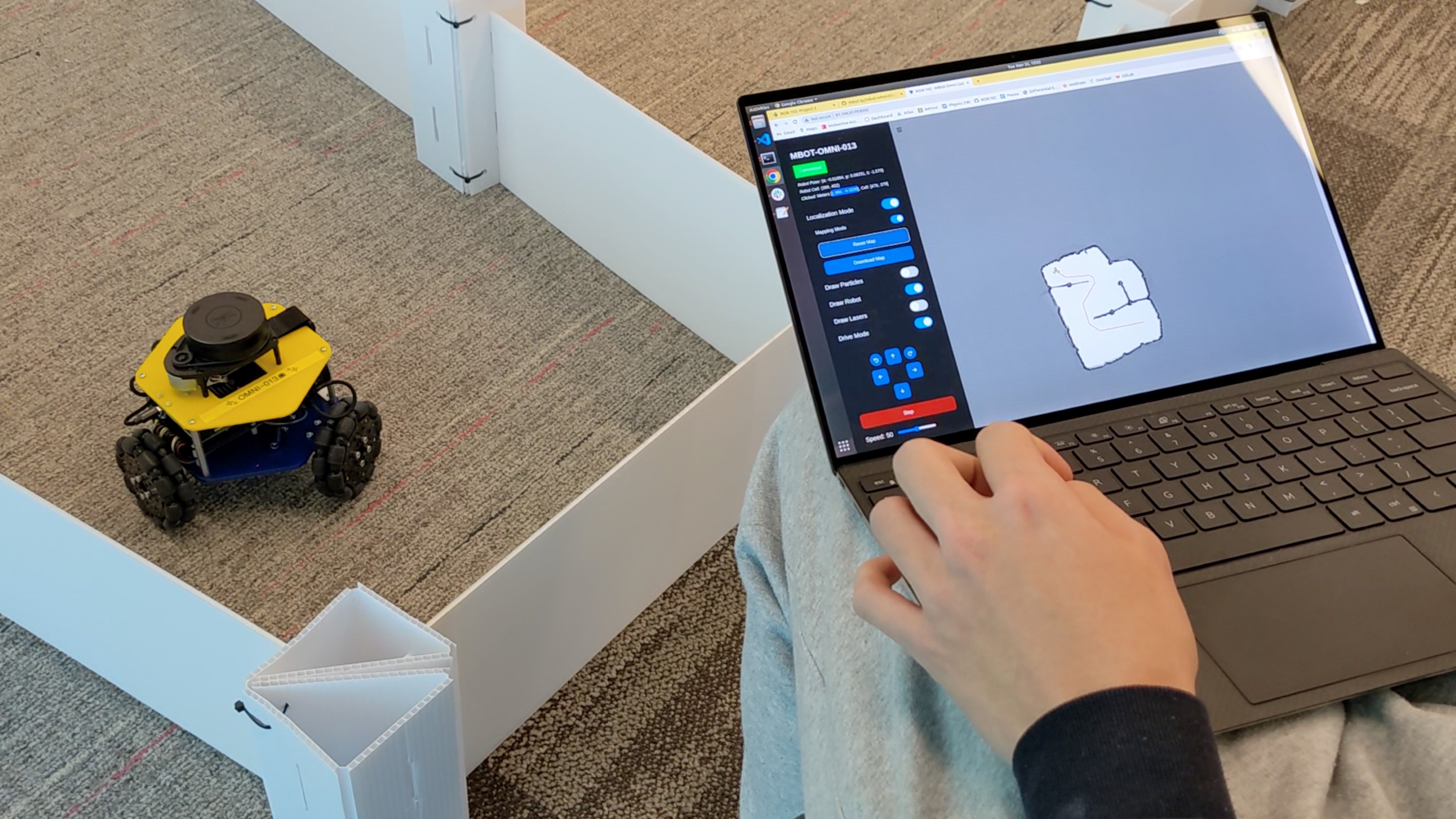}
    \caption{The MBot Web App in action. The web app can be accessed through a browser on any device connected to the robot's network, and enables teleoperation of the robot, control over the SLAM state, and visualization.}
    \label{fig:webapp}
\end{figure}

Core to the philosophy behind the MBot system is that the robots should be user-friendly from the perspective of a typical undergraduate student. The MBot Web App provides visualization and control of the platforms. The app is hosted directly on the robot's signle-board computer, and can be accessed from a browser from any device (e.g. personal computer or cell phone) connected to the same network as the robot. 
As such, interfacing with the robot through the web app requires no installation or technical prerequisites once installed on the robot, unlike other common visualization tools such as RViz. 

The web app includes a driving interface for sending robot velocity commands either through a joystick or keyboard control. It also allows the user to toggle between idle mode, localization mode, or full SLAM (mapping and localization) mode, as well as reset the map. The web app displays the current map, the robot position in the map, the laser scan, and any published paths. 
A visualization of the web app is shown in Figure~\ref{fig:webapp}.

\section{Teaching with the MBot}\label{sec:teaching}

\begin{table*}[t]
\centering
\caption{\label{tab:mbot_courses} Courses taught with the MBot Platform.}
\renewcommand{\arraystretch}{1.1}
\small
\begin{tabular}{r c c c l}
\toprule
Course & Institution & Years Offered & Student Level & Course Content \\
\midrule
Robotics Systems Lab & UM & 2014 -- Pres. & Grad. & Control systems (BalanceBot), localization \& \\
 & & & & mapping, path planning \\
Autonomous Robotics & UM & 2014 -- Pres. & Undergrad. & Localization \& mapping, path planning,  \\
 & & & & independent design project \\
SLAM \& Navigation & UM & 2022 -- Pres. & Undergrad. & Localization \& mapping, path planning \\
Intro to AI \& Programming & UM & 2021 -- Pres. & Undergrad. & Intro. to C++ \& Python, feedback control, \\
 & & & & autonomous navigation, image classification \\
Intro to AI \& Autonomous Systems & BC & 2021, 2023 & Undergrad. & Feedback control, autonomous navigation,  \\
 & & & & image classification \\
Robotics: Autonomous Navigation & HU & 2023 & Undergrad. & Intro. to programming, wall following, \\
 & & & & autonomous navigation \\ 
\midrule
\multicolumn{5}{l}{UM = University of Michigan; BC = Berea College; HU = Howard University} \\
\bottomrule
\end{tabular}
\end{table*}

A summary of the courses which have been taught by the MBot thus far can be found in Table~\ref{tab:mbot_courses}. Below, we highlight two flagship courses based around the MBot for advanced (graduate or senior undergraduate) and beginner (first year undergraduate) students.

\subsubsection{Robotics Systems Lab}
The Robotics Systems Lab is one of the core classes in the University of Michigan Robotics graduate program. The course utilizes the MBot extensively in order to address several learning objectives, most notable those surrounding statistical inference, controls, and trajectory planning. Students implement a particle filter based Markov Random Field SLAM solution utilizing the onboard LiDAR and odometry sensors in order to autonomously explore, then subsequently navigate, a previously unseen labyrinth. For this class, students are exposed to the full breadth of the the MBot software stack. Software is written in C/C++ using the Lightweight Communications and Marshalling (LCM)~\cite{huang2010lcm} messaging framework. The low-level command and control code bases are fully available for students to implement or modify as necessary. The configuration used is shown in the bottom row in Figure~\ref{fig:teaching_arch}.
As of this writing, 722 students have successfully completed the Robotics Systems Lab course, more than 50 of whom have done so as hybrid or fully remote students.

\subsubsection{Hello, Robot! Introduction to Robotics and AI}
As part of the University of Michigan's Robotics undergraduate degree, an introductory programming class based on the MBot for first-year students was developed. The course has no prerequisites and covers introductory programming through projects on the MBot Omni, including wall following, bug navigation, path planning, and image classification. The course uses the MBot Bridge API as a simple interface to the autonomy software. 
The configuration used is shown in the middle row in Figure~\ref{fig:teaching_arch}.
This course has been adapted and offered at Berea College, Howard University, and Morehouse College.%
\footnote{Details of the course \textit{Introduction to AI and Programming} are available at: \href{https://hellorob.org/}{https://hellorob.org/}.}

\section{Related Work}\label{sec:comparison}

The study of robotics and education has a long history of scientific exploration.  This history dates back to the seminal work of Papert~\cite{papert1980mindstorms} in his book {\em Mindstorms}.  This work established the foundational value proposition for computational literacy as a critical need for society.  Through the invention of the LOGO Programming Language, Papert et al.~\cite{solomon2020history} emphasized the importance of learning computer programming through robotics, where through ``embodiment as the physical computer, computation opens a vast universe of things to do.''  

Research into robotics education has grown rapidly since these early beginnings, especially for undergraduate and graduate education.  We refer the reader to the survey by Miller and Nourbakhsh~\cite{miller2016robotics} for an excellent overview of robotics education up through the mid-2000s.  There is a wealth of studies into methods and efficacy for robotics education across various levels of education~\cite{berry2016robotics, anderson2011affecting}, and its effects and potential benefits for student diversity~\cite{gini2006using}.  With their groundbreaking Robotics Engineering major at Worcester Polytechnic Institute (WPI), Gennert et al.~\cite{gennert2018robotics} made the first major steps into explorations creating a whole curriculum around this discipline of robotics.  Our design of the MBot and the Michigan Robotics curriculum drew considerable inspiration and insight from the Robotics Engineering program at WPI.

The era of scalable and programmable robot platforms was catalyzed by Martin's {\em Robotics Explorations} book~\cite{martin2000robotic} and the introduction of the LEGO Mindstorms, borrowing from the spirit from Papert et al.  Following the LEGO Mindstorms were a number of highly impactful mobile robot platforms for education.  These platforms included the Parallax Scribbler~\cite{summet2009personalizing}, Sony AIBO~\cite{veloso2006cmrobobits}, and modified versions of the iRobot Roomba~\cite{dickinson2007roomba, conbere2007toys, lapping2008wiimote, tribelhorn2007evaluating}, as forerunners to the Willow Garage Turtlebot~\cite{amsters2020turtlebot}.  Robotics education also saw the development of its own robot programming environments, such as Tekkotsu~\cite{touretzky2005tekkotsu}, more amenable to undergraduate teaching than common robot middleware frameworks.

Robots available today for higher education tend to fall into two categories. There are those where the focus is on and interfacing of basic sensors and the low level control of motors so that the user may program the robot to perform simple tasks like line following.  These platforms, like VEX~\cite{vex}, the Pololu 3Pi+~\cite{3pi} or the SparkFun RedBot~\cite{redbot}, usually provide a simple way to program the robots, like with a custom library in the Arduino IDE.  There is great educational value in working with a system like this to learn about programming embedded systems, but the sophistication of behaviours students are able to program into the robot is typically limited.  These robots are therefore best suited to high school or lower level undergraduate education as a part of a beginner curriculum.   

On the other end of the spectrum are robots like Turtlebot~\cite{turtlebot}, Duckiebot~\cite{paull2017duckietown}, JetBot~\cite{jetbot},  MIT RaceCar~\cite{mitracecar}, and MuSHR~\cite{srinivasa2019mushr}, which are capable of interfacing with multiple high bandwidth sensors, cameras and Lidar, and utilize mature and sophisticated robotics software and algorithms in ROS.  These robots are typically suited for students and researchers with a strong background in computers and programming as part of an advanced curriculum or in research.  The MBot was designed to meet needs across undergraduate and graduate levels of education and be accessible and adaptable for many types of institutions, students, and courses.

\begin{table}[t]
\centering
\caption{\label{tab:robot_comparison} Comparison of available robot platforms}
\renewcommand{\arraystretch}{1.1}
\small
\begin{tabular}{r c c l}
\toprule
                    &        & Camera or &  \\
Platform            & Cost   & LiDAR   &  Drive type \\
\midrule
Pololu 3Pi+         & \$150  & -- & DD  \\
SparkFun RedBot     & \$140  & -- & DD  \\
Parallax Scribbler  & \$180  & -- & DD\\ 
Vex V5 Kit          & \$750  & -- & Multi\\ 
DuckieBot           & \$450  & Camera & DD\\ 
JetBot              & \$260  & Camera & DD\\  
TurtleBot 4 Lite    & \$1200 & Both & DD\\ 
MuSHR               & \$1525 & Both & Ack\\ 
MIT Racecar         & \$2600 & Both & Ack\\ 
AgileX Limo         & \$2900 & Both & Multi\\ 
\midrule
\textbf{MBot (Ours)}&\$100-\$400 & Both  & Multi\\
\midrule
\multicolumn{4}{l}{DD = Differential; Ack = Ackermann; Multi = Reconfigurable} \\
\bottomrule
\end{tabular}
\end{table}

\section{Conclusion}

We present the MBot ecosystem for robotics education at the undergraduate and graduate levels. The platform has been developed over nearly a decade of robotics education at the University of Michigan. It has proven successful at a number of courses across multiple institutions. 
Future work on the platform will involve working with partnering institutions to integrate it into more courses, and further expanding the suite of tools available. 

\section{Acknowledgments}

This work was supported in part by Ford Motor Company, J.P.~Morgan AI Research, Amazon Inc., Toyota Research Institute, the Sloan Foundation, an NSERC scholarship, and a generous donation from Roger Ehrenberg and Carin Levine-Ehrenberg.
Many of the photos and videos of the MBot were taken and edited by Dan Newman, who we thank for lending his effort and talent to this work.
The MBot Team is also deeply grateful to Jessy Grizzle, whose forward-thinking leadership of Michigan Robotics enabled the MBot community and our undergraduate Robotics curriculum to grow from a spark of ingenuity into a thriving ecosystem of innovation in robotics.

This MBot, its evolution, and its service to students and courses at Michigan Robotics is indebted to the work and collaboration of many outstanding contributors, including: 
Alphonsus Adu-Bredu, Bahaa Aldeeb, Senthur Raj Ayyappan, Onur Bagoren, Carlotta Berry, Xiaotong Chen, Shane DeMeulenaere, Xiaoxiao Du, Ryan Eustice, Daniel Fairfax, Paul Foster, Bryar Frank, Kevin French, Michael Gonzalez, Mark Guzdial, Collin Johnson, Jasmine Jones, Dwayne Joseph, Zach Kaufman, Cameron Kisailus, Ben Kuipers, Chien Erh Lin, Chin-Wei Lin, Havel Liu, Isaac Madhavaram, Yves Nazon, Edwin Olson, Elizabeth Olson, Jan Pearce, Broderick Riopelle, Ali Robinson, Elliott Rouse, Priscilla Saarah, Stephen Seymour, Todd Shurn, Katie Skinner, Zhiqiang Sui, Joseph Taylor, Justin Tesmer, Grey Thomas, Maxwell Topping, Jon Toto, Karthik Urs, Sophie van Genderen, Jorge Vilchis, Franklin Volcic, Wei Wu, Ruihan Xu, Zhen Zeng, and Zheming Zhou.

\bibliographystyle{IEEEtran}
\bibliography{ref.bib}

\end{document}